\documentclass[conference]{IEEEtran}
\IEEEoverridecommandlockouts
\usepackage{cite}
\usepackage{amsmath,amssymb,amsfonts}
\usepackage{algorithmic}
\usepackage{graphicx}
\usepackage{textcomp}
\usepackage{xcolor}
\def\BibTeX{{\rm B\kern-.05em{\sc i\kern-.025em b}\kern-.08em
    T\kern-.1667em\lower.7ex\hbox{E}\kern-.125emX}}
\newtheorem{lemma}{Lemma}
\newtheorem{proposition}{Proposition}

\newtheorem{definition}{Definition}
\newtheorem{theorem}{Theorem}

\begin{document}

\title{
 Online Learning from Strategic Human Feedback in LLM Fine-Tuning  

\thanks{This work has been accepted to be presented at IEEE ICASSP'25.}
\thanks{This work is supported in part by the Ministry of Education, Singapore, under its Academic Research Fund Tier 2 Grant with Award no. MOE-T2EP20121-0001; in part by SUTD Kickstarter Initiative (SKI) Grant with no. SKI 2021\_04\_07; and in part by the Joint SMU-SUTD Grant with no. 22-LKCSB-SMU-053.}
% Incentive-Compatible Mechanism Design for Aggregating Online Human Feedback in LLM Fine-Tuning
 }

\author{\IEEEauthorblockN{Shugang Hao, Lingjie Duan}
\IEEEauthorblockA{\textit{Pillar of Engineering Systems and Design, Singapore University of Technology and Design} \\
shugang\_hao@sutd.edu.sg, lingjie\_duan@sutd.edu.sg}
% \and
% \IEEEauthorblockN{Lingjie Duan}
% \IEEEauthorblockA{\textit{Pillar of Engineering Systems and Design} \\
% \textit{Singapore University of Technology and Design}\\
% Singapore, Singapore \\
% lingjie\_duan@sutd.edu.sg}
}

\maketitle

\begin{abstract}
Reinforcement learning from human feedback (RLHF) has become an essential step in fine-tuning large language models (LLMs) to align them with human preferences. However, human labelers are selfish and have diverse preferences. They may strategically misreport their online feedback to influence the system’s aggregation towards their own preferences. Current practice simply averages labelers' feedback per time and fails to identify the most accurate human labeler, leading to linear regret $\mathcal{O}(T)$ for $T$ time slots. To our best knowledge, we are the first to study online learning mechanisms against strategic human
labelers in the LLM fine-tuning process. We formulate a new dynamic Bayesian game and dynamically adjust human labelers' weights in the preference aggregation, ensuring their truhtful feedback and sublinear regret $\mathcal{O}(T^{1/2})$. Simulation results demonstrate our mechanism's great advantages over the existing benchmark schemes. 
\end{abstract}


\section{Introduction}\label{S1}

Large language models (LLMs) such as ChatGPT and SORA have succeeded in handling a number of tasks such as text and video generation. To better meet users' demands for specific applications, pre-trained LLMs are fine-tuned to be customized using task-oriented datasets (e.g., \cite{touvron2023open}). Traditional supervised learning methods fail to align with human preferences because of the difficulty in acquiring a significant number of question-answer paired data (e.g., \cite{kopf2024openassistant}, \cite{sun2024mechanism}). Reinforcement learning from human feedback (RLHF) has emerged as a promising approach to tackle this human preference alignment problem (e.g., \cite{christiano2017deep}, \cite{ouyang2022training}). It queries online human feedback (e.g.,  in a weekly cadence \cite{bai2022training}, \cite{touvron2023llama}, \cite{xiong2024iterative}) to obtain a human-annotated preference dataset, which will then be used to train and update the learning policy. RLHF has become an essential training step in LLM fine-tuning due to its effectiveness in aligning with human preferences.

 However, human labelers are selfish in the RLHF loop to have diverse preferences and they may strategically misreport their online feedback to influence the system's aggregation towards their own preferences (e.g., \cite{sun2024mechanism},\cite{soumalias2024truthful}). For example, a user in an LLM rating system may strategically give an extreme response rating of 0 or 10 in the range of [0, 10] to maximally influence the overall rating toward his actual rate (e.g., \cite{conitzer2024social}). Besides, there is a renowned ``wet bias" where a weather forecaster as human labeler or predictor may deliberately report an exaggerated probability of precipitation to increase the influence of his forecast in the system's final prediction (e.g., \cite{roughgarden2017online}). Current practice of LLM fine-tuning largely ignores human labelers' misreporting and   
 simply averages human feedback in the preference aggregation (e.g., \cite{christiano2017deep}, \cite{ouyang2022training}, \cite{rafailov2024direct}, \cite{xie2024exploratory}, \cite{cen2024value}, \cite{zhang2024self}, \cite{chen2024optune}) and we wonder its actual performance. Our first question naturally arises:
\begin{itemize}
    \item \textit{Q1. How bad is the current practice of average feedback aggregation for LLM fine-tuning performance?}
\end{itemize}

Later we prove that the average feedback aggregation scheme fails to identify the most accurate human labeler in the
online learning process and incurs a non-vanishing regret $\mathcal{O}(T)$ overtime. This motivates us to propose new schemes for truthful human feedback and vanishing regret. In the recent RLHF literature, Sun \textit{et al.} (2024) in \cite{sun2024mechanism}, Park \textit{et al.} (2024) in \cite{park2024rlhf}, Soumalias \textit{et al.} (2024) in \cite{soumalias2024truthful} and Dubey \textit{et al.} (2024) in \cite{dubey2024auctions} focus on monetary payment-based mechanism design to reward and enable strategic human labelers' truthful preference feedback. In practice, monetary mechanisms involve complicated billing issues and may not be easy to implement.  Furthermore, these works assume a one-shot or offline preference feedback setting and do not consider human labelers' online feedback. In the online setting, human labelers have more room to strategically misreport and play with the RLHF system for long-term influence. 

In the related literature of  algorithmic game theory, there are relevant non-monetary mechanism studies on facility location games (e.g., \cite{asadi2022collaborative}, \cite{chen2024mechanism}, \cite{li2024strategyproof}), where the system aims to incentivize customers' truthful reporting of their locations to optimize facility placement. There each customer can strategically misreport his location to mislead the facility placement as close to his location (preference) as possible. The popular ``median" scheme (e.g., \cite{conitzer2024social}, \cite{wang2024positive}) to aggregate multi-agent reports  is widely used to return customers' truthful reporting. Yet, later we prove that it is no longer truthful for our online problem and also incurs a non-vanishing regret as the average feedback aggregation scheme. Our second question is thus: 
\begin{itemize}
    \item \textit{Q2. How to design an efficient and truthful mechanism for online learning from strategic human
feedback in LLM fine-tuning? }
\end{itemize}

A natural non-monetary mechanism  approach is to deploy a weighted aggregation and dynamically adjust each human labeler's weight in the online RLHF process. However, it is challenging to motivate human labelers' truthful preference feedback for a vanishing regret. Human labelers' preferences are hidden and can vary over time slots, which makes it difficult for the system to verify and correct their misreports to learn their preferences (e.g., \cite{sun2024mechanism}). Since the most accurate human labeler is unknown in the online learning process, it is difficult for the system to dynamically weigh each human labeler to achieve a vanishing regret.

To our best knowledge, we are the first to study online learning mechanisms against
strategic human labelers in LLM fine-tuning. We
formulate a new dynamic Bayesian game and dynamically adjust each human labeler's weight in the aggregation according to their feedback accuracy, which ensures their truthful feedback and sublinear regret $\mathcal{O}(T^{1/2})$. Finally, simulation results also demonstrate our mechanism’s great advantages over the existing benchmark schemes.

\section{System Model and Problem Formulation}  \label{S2}

First, we introduce our system model. Then, we formulate a new dynamic Bayesian game and give desired properties for guiding our late mechanism design.

\subsection{System Model of LLM Fine-Tuning from Online Human Feedback}\label{S2.1}

We consider an LLM fine-tuning process of aggregating online feedback from $N$$\geq$2 strategic human labelers overtime. It starts from fine-tuning a pre-trained language model with supervised learning based on a task-orientated dataset (e.g., paragraph summary) to obtain a reference policy $\pi_{\texttt{ref}}$. Then, the system iterates the RLHF process in $T$ time slots (e.g., a weekly cadence \cite{bai2022training}, \cite{touvron2023llama}, \cite{xiong2024iterative}), where each time slot $t \in [T] := \{1, \cdots, T\}$ contains the following three stages. The current practice of LLM fine-tuning only involves Stages I and II with uniform weights in the aggregation and we introduce Stage III to dynamically adjust human labelers' weights according to their online feedback accuracy.

\textit{Stage I.} \textit{Online Preference Feedback}: The system draws $m_t$ prompts $\{x_j^{t}\}_{j=1}^{m_t}$ from the context space $\mathcal{X}$ and generates $m_t$ pairwise responses $\{(y_{l_j}^{t}, y_{l_j'}^{t}|x_j^{t})\}_{j=1}^{m_t}$ from the response space $\mathcal{Y}$ according to the last slot policy $\pi_{t-1}$ with $\pi_{0} = \pi_{\text{ref}}$ (e.g., \cite{xiong2024iterative}). It then shares 
$\{(x_j^t, y_{l_j}^{t}, y_{l_j'}^{t})\}_{j=1}^{m_t}$ with $N$ human labelers for their preference feedback. Each human labeler $i$$\in$$[N]$ independently realizes his continuous private preference of response $y_{l_j}^{t}$ over $y_{l_j'}^{t}$ as $\mathcal{P}_i(y_{l_j}^{t} \succ y_{l_j'}^{t}|x_j^t) \in [0, 1]$ for each $j \in [m_t]$ and he believes that the ground-truth preference $p_j^t\sim Bernoulli(\mathcal{P}_i(y_{l_j}^{t} \succ y_{l_j'}^{t}|x_j^t))$, where realization $p_j^t = 1$ means response $y_{l_j}^{t}$ is preferred over $y_{l_j'}^{t}$ and $p_j^t = 0$ otherwise.

{He wants to influence the system’s aggregation toward his own preference} and may feedback another continuous $\mathcal{\hat{P}}_i(y_{l_j}^{t} \succ y_{l_j'}^{t}|x_j^t) \in [0, 1]$ different from his actual $\mathcal{P}_i(y_{l_j}^{t} \succ y_{l_j'}^{t}|x_j^t)$ to the system (e.g., \cite{sun2024mechanism},\cite{soumalias2024truthful}). The system and other human labelers are uncertain of his $\mathcal{P}_i(y_{l_j}^{t} \succ y_{l_j'}^{t}|x_j^t)$ realization. 

\textit{Stage II.} \textit{Online Feedback Aggregation and Policy Optimization}: After receiving each human labeler $i$'s feedback $\{\mathcal{\hat{P}}_i(y_{l_j}^{t} \succ y_{l_j'}^{t}|x_j^t)\}_{j=1}^{m_t}$ for $i \in [N]$, the system aggregates according to the weight $w_i^{t}$ for each prompt $j \in[m_t]$ as 
{
    \begin{align}
        \mathcal{P}(y_{l_j}^{t} \succ y_{l_j'}^{t}|x_j^t)=  \frac{\sum_{i=1}^N w_i^{t}\mathcal{\hat{P}}_i(y_{l_j}^{t} \succ y_{l_j'}^{t}|x_j^t)}{\sum_{i'=1}^N w_{i'}^{t}},\label{eq0}
\end{align}
}

\noindent with a uniform weight $w_i^1$=1 for all $i$$\in$$[N]$ in the first time slot.\footnote{One can also use $w_i^1$=$1/N$ without any change for the aggregation result.}  
      The aggregated preference $\{\mathcal{P}(y_{l_j}^{t} \succ y_{l_j'}^{t}|x_j^t)\}_{j=1}^{m_t}$ will be included to construct the human preference dataset $\mathcal{D}_t := \{\mathcal{P}(y_{l_j}^{t} \succ y_{l_j'}^{t}|x_j^t)\}_{j=1}^{m_t}$. Based on the preference dataset $\mathcal{D}_t$, the system then learns a policy $\pi_t$ using direct preference optimization (DPO) to solve a KL-regularized optimization problem against the reference policy $\pi_{\text{ref}}$ (e.g.,   \cite{rafailov2024direct}):
      {
      \begin{align*}
        \min_{\pi_t} -\mathbb{E}_{(x, y, y')\sim \mathcal{D}_t}  \ln \sigma\bigg( 
 \beta \ln \frac{\pi_t(y|x)}{\pi_{\texttt{ref}}(y|x)} - \beta \ln \frac{\pi_t(y'|x)}{\pi_{\texttt{ref}}(y'|x)}\bigg),
    \end{align*}
    }

    \noindent where $\sigma(\cdot)$ denotes the logistic function and $\beta$ is a parameter of evaluating the deviation from the reference policy $\pi_{\texttt{ref}}$.

 \textit{Stage III.} \textit{Reweighing Human Labelers}: 
 In the RLHF literature, the current practice of LLM fine-tuning
simply averages human labelers’ feedback in the preference
aggregation using uniform $w_i^t$=1 for all $i$$\in$$[N]$ and $t$$\in$$[T]$ in \eqref{eq0} (e.g., \cite{christiano2017deep}, \cite{ouyang2022training}, \cite{rafailov2024direct}). To our best knowledge, we are the first to consider dynamic adjustment of each human labeler's weight in the online RLHF process, where the system determines each  
\begin{align}\label{eq}
    w_i^{t+1}=f_i(\{\{\mathcal{\hat{P}}_i(y_{l_j}^{t} \succ y_{l_j'}^{t}|x_j^t)\}_{j=1}^{m_t}\}_{i=1}^N, \{p_j^t\}_{j=1}^{m_t})
\end{align}
for the next slot $t+1$'s aggregation according to feedback $\{\{\mathcal{\hat{P}}_i(y_{l_j}^{t} $$\succ$$ y_{l_j'}^{t}|x_j^t)\}_{j=1}^{m_t}\}_{i=1}^N$ and the realized preference $\{p_j^t\}_{j=1}^{m_t}$. The system implements and tests the obtained policy $\pi_t$ for customers' practical usage and learns the realized binary preference $p_j^t \in \{1, 0\}$ for each prompt $j \in [m_t]$ according to its customers' realized experience, where $p_j^t=1$ if $y_{l_j}^{t}$ is preferred than $y_{l_j'}^{t}$ and 0 otherwise (e.g., \cite{pacchiano2021dueling}, \cite{chen2022human}, \cite{zhong2024dpo}).

Each selfish human labeler wants to use $\mathcal{\hat{P}}_i(y_{l_j}^{t} \succ y_{l_j'}^{t}|x_j^t)$ to influence the system’s aggregation towards his own preference (e.g., \cite{sun2024mechanism}, \cite{soumalias2024truthful}). Thus, he wants to obtain a large weight in the system's feedback aggregation in \eqref{eq0} in each time slot $t \in [T]$ and
 aims to maximize his long-term influence benefit as the cumulative weight over the whole $T$ time slots as follows:
 {
\begin{align}\label{e2}
&u_i(\{\{\mathcal{\hat{P}}_i(y_{l_j}^{t} \succ y_{l_j'}^{t}|x_j^t)\}_{j=1}^{m_t}\}_{t=1}^T) \\ 
=&\sum_{t=1}^{T}w_i^{t}(\{\{\mathcal{\hat{P}}_i(y_{w_j}^{t-1} \succ y_{l_j'}^{t}|x_j^{t-1})\}_{j=1}^{m_{t-1}}\}_{i=1}^N, \{p_j^{t-1}\}_{j=1}^{m_{t-1}}). \nonumber
\end{align}
}

On the other hand, the system wants to improve the feedback accuracy in the aggregation by assigning the largest weight to the most accurate human feedback. However, the best  human labeler is unknown in the online iteration. It then turns to reducing the regret between online weighted aggregation and offline choice of the best human labeler in hindsight (e.g., \cite{roughgarden2017online}, \cite{frongillo2021efficient}, \cite{freeman2020no}), where the performance loss or online regret is defined as the mean square error (MSE) between the system's weighted aggregation in \eqref{eq0} and the realized binary preference as follows:
\begin{align}
    R(T) :=& \sum_{t=1}^T\frac{1}{m_t}\sum_{j=1}^{m_t}\bigg(\sum_{i=1}^N  
\frac{w_i^{t}\mathcal{\hat{P}}_i(y_{l_j}^{t} \succ y_{l_j'}^{t}|x_j^t)}{\sum_{i'=1}^N w_{i'}^{t}} - p_{j}^t\bigg)^2 \nonumber \\
% \end{align}
% \begin{align}
    &- \min_{i \in [N]} \sum_{t=1}^T\frac{1}{m_t}\sum_{j=1}^{m_t}\big(\mathcal{P}_i(y_{l_j}^{t} \succ y_{l_j'}^{t}|x_j^t) - p_{j}^t\big)^2. \label{e3}
\end{align}
% where the system wants to find the most accurate human labeler for the efficiency of the RLHF learning.

\subsection{Dynamic Bayesian Game Formulation}\label{S2.2}

Based on our system model above,  we formulate the multi-agent online learning as a new dynamic Bayesian game:
\begin{itemize}
    \item In Stage I of each time slot $t$$\in$$[T]$, each human labler $i$ with his private preference $\{\mathcal{P}_i(y_{l_j}^{t}$$\succ$$y_{l_j'}^{t}|x_j^t)\}_{j=1}^{m_t}$ determines his preference feedback $\{\mathcal{\hat{P}}_i(y_{l_j}^{t}$$\succ$$y_{l_j'}^{t}|x_j^t)\}_{j=1}^{m_t}$ to maximize his accumulative weight in \eqref{e2}.
    \item In Stage III of each time slot $t$$\in$$[T]$, the system updates each human labeler's weight $w_i^{t+1}$=$f_i(\{\{\mathcal{\hat{P}}_i(y_{l_j}^{t} \succ y_{l_j'}^{t}|x_j^t)\}_{j=1}^{m_t}\}_{i=1}^N, \{p_j^t\}_{j=1}^{m_t})$ for reducing regret in \eqref{e3}.
\end{itemize}

% Since human labelers’ preferences are hidden
% and vary over time slots, the system finds it difficult to verify
% and correct their misreports to learn their preferences (e.g.,
% [3]). Further, human labelers can strategically manipulate their
% feedback from time to time against the system’s learning or
% inference, making it even harder for the system to learn the
% ground truth for weight update.

% To handle the above two challenges, 
Note that there is no strategic decision for human labelers or the system
in Stage II. We need to carefully design an online aggregation mechanism for ensuring each human labeler's truthful preference feedback and a vanishing regret in time. We define the desired properties as below.

\begin{definition}[Truthfulness for Human Feedback]\label{def1}
An online weighted aggregation mechanism $\mathcal{M}$ is truthful if each human labeler $i$$\in$$[N]$ obtains a larger long-term influence in \eqref{e2} over the whole $T$ time slots though truthful preference feedback instead of misreporting in the mean time, i.e., 
\begin{align*}
    \mathbb{E} \bigg[\sum_{t=2}^{T-1} w_{i}^{t+1} \big(&\{\mathcal{P}_i(y_{l_j}^{t} \succ y_{l_j'}^{t}|x_j^t)\}_{j=1}^{m_t}, \{p_{j}^t\}_{j=1}^{m_t}, \\
    &\{\{\mathcal{\hat{P}}_k(y_{l_j}^{t} \succ y_{l_j'}^{t}|x_j^t)\}_{j=1}^{m_t}\}_{k=1, k\ne i}^N\big)\bigg] \\
    \geq \mathbb{E} \bigg[\sum_{t=2}^{T-1} w_{i}^{t+1} \big(&\{\mathcal{\hat{P}}_i(y_{l_j}^{t} \succ y_{l_j'}^{t}|x_j^t)\}_{j=1}^{m_t}, \{p_{j}^t\}_{j=1}^{m_t}, \\
   & \{\{\mathcal{\hat{P}}_k(y_{l_j}^{t} \succ y_{l_j'}^{t}|x_j^t)\}_{j=1}^{m_t}\}_{k=1, k\ne i}^N\big)\bigg].
\end{align*}
\end{definition}

% Further, we expect the final regret $R_T$ to be sub-linear in $T$ as the performance guarantee for the system's aggregation. We define below.
\begin{definition}[High Efficiency in Sublinear Regret $R(T)$ in \eqref{e3}]
    An online weighted aggregation mechanism $\mathcal{M}$ is efficient if its time-average regret $R_\mathcal{M}(T)/T$ is vanishing in the time slot number $T$, i.e., $\lim_{T\to\infty} 
\frac{R_\mathcal{M}(T)}{T} = 0.$
\end{definition}

% Due to the page limit, we skip some detailed proofs but still introduce the main ideas and intuitions in the main paper. Please refer to our online technical report \cite{hao2024DML} for details.

\section{Two Benchmark Schemes: Average Feedback Aggregation and Median Aggregation}\label{S3}

In this section, we analyze two common schemes used in the literature of both RLHF and algorithmic game theory, serving as two fair benchmarks for our  mechanism to compare later.

\subsection{Benchmark 1: Average Feedback Aggregation}

The current practice of LLM fine-tuning simply averages human labelers’ feedback in the preference
aggregation using uniform $w_i^t = 1$ for all $i \in [N]$ and $t \in [T]$ (e.g., \cite{christiano2017deep}, \cite{ouyang2022training}, \cite{rafailov2024direct}). 
% Since each human labeler's weight is fixed and independent of his feedback accuracy, he may feedback an arbitrary $\mathcal{\hat{P}}_k(y_{l_j}^{t} \succ y_{l_j'}^{t}|x_j^t)$ to the system. 
Unfortunately, such an average feedback aggregation scheme can lead to a non-vanishing regret as shown below.
\begin{lemma}\label{L1}
   The system's regret in \eqref{e3} under the benchmark 1 of average aggregation is $R_1(T)$=$\mathcal{O}(T)$, leading to a non-vanishing time-average regret $\lim_{T\to\infty}\frac{R_1(T)}{T}> 0$. 
\end{lemma}

Benchmark 1 fails to dynamically adjust human labelers' aggregation weights according to their online feedback accuracy. Thus, the most accurate human labeler cannot receive the largest weight in the online learning process, leading to a non-vanishing time-average regret even if $T \to \infty$. We are motivated to find other schemes to reduce the system’s regret.  

\subsection{Benchmark 2: Median Aggregation Scheme}\label{S3.2}

In the algorithmic game theory literature, the popular ``median" scheme is widely used to motivate strategic agents' truthful reporting (e.g., \cite{conitzer2024social}, \cite{wang2024positive}).  We define it below.

\begin{definition}[Median Aggregation Scheme]\label{defm} 
The system first re-organizes human labelers' preference feedback $\{\mathcal{\hat{P}}_i(y_{l_j}^{t} \succ y_{l_j'}^{t}|x_j^t)\}_{i=1}^N$ in an increasing order as $\mathcal{\hat{P}}_{k_1,j}^t \leq \cdots \leq \mathcal{\hat{P}}_{k_N,j}^t$ for each prompt $j \in [m_t]$ in each time slot $t \in [T]$. It then chooses the median $\mathcal{\hat{P}}_{k_s,j}^t$ as its preference aggregation, where the index $s = N/2$ if $N$ is even and $s=(N+1)/2$ otherwise.
% Given human labelers' preference feedback $\{\mathcal{\hat{P}}_i(y_{l_j}^{t} \succ y_{l_j'}^{t}|x_j^t)\}_{i=1}^N$ for each prompt $j \in [m_t]$ in each time slot $t \in [T]$, the system  chooses the median feedback with index $m_j \in [N]$ as its final decision, i.e., $w_{m_j}^t = 1$ and $w_{i_j}^t = 0$ for $i \ne m$ and $i \in [N]$.
\end{definition}

Since the benchmark 2 independently commits to the median feedback for each prompt in each time slot, at the equilibrium, all the human labelers' feedback will converge to one common point for an equal probability to be the median. We summarize the equilibrium in the following.
\begin{proposition}\label{L0}
At the equilibrium of the benchmark 2, all the human labelers' feedback $\mathcal{\hat{P}}_i^*(y_{l_j}^{t} \succ y_{l_j'}^{t}|x_j^t)$ will converge to an arbitrary point $ \mathcal{\hat{P}}(y_{l_j}^{t}$$\succ $$y_{l_j'}^{t}|x_j^t)$$\in$$[0, 1]$ for $j$$\in$$[m_t]$, $i$$\in$$[N]$ and $t$$\in$$[T]$, which may not be the same as their own preference.   
\end{proposition}

The median scheme is not truthful since a human labeler may be committed with a positive probability by misreporting than no probability by truthful feedback. Further, it leads to a non-vanishing time-average regret in $T$ in the following.
\begin{lemma}\label{L1-1}
   The system's regret in \eqref{e3} under the untruthful median scheme is $R_2(T) = \mathcal{O}(T)$, leading to a non-vanishing time-average regret $\lim_{T\to\infty}\frac{R_2(T)}{T}> 0$.
\end{lemma}

The median feedback can lead to a total aggregation loss of $\mathcal{O}(T)$ over the $T$ time slots. Yet, there can exist a human labeler holding $\mathcal{P}_k(y_{l_j}^{t} \succ y_{l_j'}^{t}|x_j^t) = p_j^t$, incurring zero aggregation loss of the best human labeler in hindsight. The average regret is then non-vanishing even if the time slot number $T\to\infty$. Given non-vanishing regret of both benchmark schemes,
 we are well motivated to develop a truthful mechanism to substantially reduce the system’s regret.

\section{Online Weighted Aggregation Mechanism: New Design, Analysis, and Simulations}\label{S4}

In this section, we first present our  mechanism design and its regret bound. We then run simulations for verification.

\subsection{Mechanism Design and Theoretical Analysis} 

 Unlike benchmark 1, in Stage III of each time slot, we dynamically adjust each human labeler's weight based on his feedback accuracy and assign a larger weight if his feedback is closer to the realized binary preference. We need to carefully design the online mechanism in \eqref{eq} to ensure that each obtains the largest long-term influence in Definition~\ref{def1} only with truthful feedback. 
We define our mechanism as below.  
\begin{definition}[Online Weighted  Aggregation Mechanism]\label{def3}
At Stage III of each time slot $t \in [T-1]$, the system updates each human labeler's weight $w_i^{t+1}$ in \eqref{eq} based on his feedback $\mathcal{\hat{P}}_i(y_{l_j}^{t} \succ y_{l_j'}^{t}|x_j^t)$ and the realized binary preference $p_j^t$: 
\begin{align}\label{e4}
    w_{i}^{t+1} = w_{i}^{t} \cdot \bigg(1 - \alpha\frac{1}{m_t}\sum_{j=1}^{m_t} \big(\mathcal{\hat{P}}_i(y_{l_j}^{t} \succ y_{l_j'}^{t}|x_j^t)- p_j^t\big)^2 \bigg),
\end{align}
where $\alpha > 0$ is the step-size parameter.
\end{definition}

Intuitively, our mechanism determines each human labeler's weight in time slot $t+1$ based on his feedback accuracy in the previous time slot $t$. If the squared difference between his feedback and the realized binary preference $(\mathcal{\hat{P}}_i(y_{l_j}^{t} \succ y_{l_j'}^{t}|x_j^t)- p_j^t)^2$ is small, his weight $w_i^{t+1}$ will be only reduced by a small value from $w_i^{t}$. Though all human labelers' weights are decreasing over time, we care about the relative weighted aggregation as in \eqref{eq0}.

Since each human labeler holds a Bernoulli belief on $p_j^t$, our mechanism satisfies the truthful property as shown below.
\begin{proposition}\label{L2}
    Our mechanism in Definition~\ref{def3} is truthful, i.e., $\mathcal{\hat{P}}_i^*(y_{l_j}^{t} \succ y_{l_j'}^{t}|x_j^t) = \mathcal{P}_i(y_{l_j}^{t} \succ y_{l_j'}^{t}|x_j^t)$ for any prompt $j \in [m_t]$, human labeler $i \in [N]$ and time slot $t \in [T]$.
\end{proposition}

Further, our mechanism is efficient and incurs a vanishing time-average regret in $T$ in the following.
\begin{theorem}\label{Thm1}
    Our mechanism in Definition~\ref{def3} incurs the sublinear regret $R_{\mathcal{M}}(T) $=$\mathcal{O}(T^{\frac{1}{2}})$ by choosing step-size $\alpha =\frac{2}{3}\sqrt{\frac{2\ln N}{T}}$, leading to zero time-average regret with $\lim_{T\to\infty} 
\frac{R_\mathcal{M}(T)}{T} = 0$.
\end{theorem}

According to Theorem~\ref{Thm1}, our mechanism obviously improves from benchmarks 1 and 2 by distinguishing the best human labeler in the online process as $T\to\infty$. As $N$ increases, the system may find a more accurate human labeler in hindsight. Thus, it chooses a larger step-size $\alpha$ to 
punish inaccurate human labelers more in the weighted aggregation to retire them. As $T$ increases, the system is more patient to choose a smaller $\alpha$ for selecting the best human labeler in hindsight with more time slots and samples. 

\subsection{Simulation Results for Verification}\label{S4.2}

In this subsection, we run simulations to show our mechanism's great improvement from the two benchmark schemes. 

Based on human-written form post
summaries with TRLX framework (e.g., \cite{havrilla2023trlx}) for the RLHF
process, we consider a task of paragraph summarization (e.g., \cite{rafailov2024direct}) and use a supervised fine-tuning model to obtain the reference policy $\pi_{\text{ref}}$. We use the Reddit TL;DR summarization dataset in \cite{volske2017tl} to draw each prompt $x_j^t$ as a form post from Reddit, and the policy $\pi_t$ generates each pairwise summary $(y_{l_j}^t, y_{l_j'}^t)$ of the main points in the post. We follow \cite{rafailov2024direct} to use DPO for the policy training. We use synthetic data to randomly generate each human labeler $i$'s preference $\mathcal{P}_i(y_{l_j}^{t} \succ y_{l_j'}^{t}|x_j^t)$ in the range of [0, 1] for each pairwise response. Further, we randomly generate the binary realized ground-truth preference $p_j^t \in \{1, 0\}$ for each prompt $j \in [m_t]$, where we fix $m_t$=50, change $N$ and $T$ for evaluation.

\begin{figure}
    \centering   \includegraphics[width=0.6\linewidth]{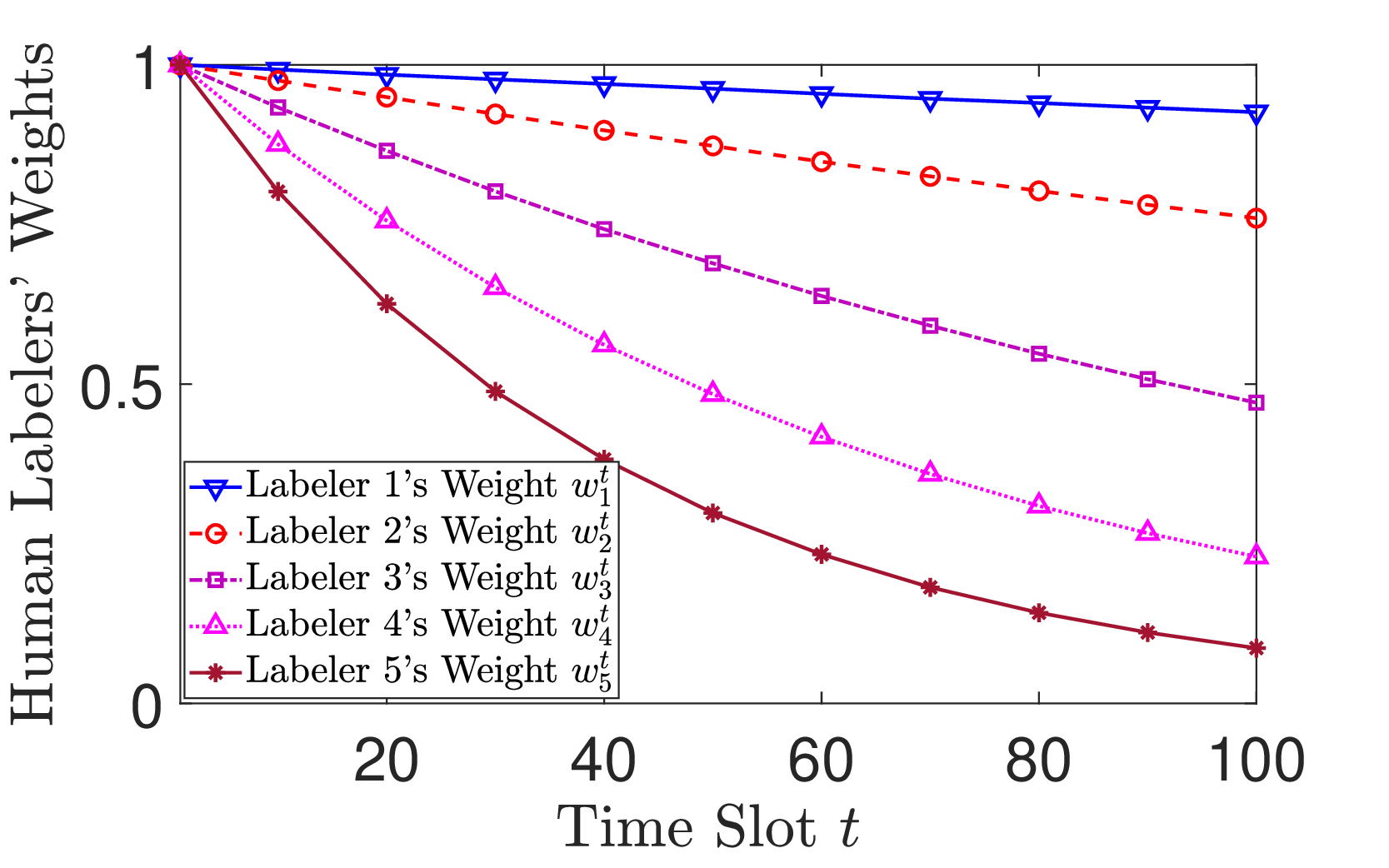}
    \caption{Each human labeler's weight $w_i^t$ versus time slot $t$. Here we fix $N=5$, $T=100$, $m_t=50$.} 
    \label{fig0}
\end{figure}

Figure~\ref{fig0} shows how each human labeler's weight $w_i^t$ evolves over time slot $t$$\in$$ [T]$. We consider human labelers 1-5 following a descent order of feedback accuracy, where labeler 1 is the most accurate and labeler 5 is the least accurate. As time evolves, our mechanism manages to assign weight proportional to any labeler's feedback accuracy. The system can thus well approximate the real preferences in the offline optimum.

% We find that our mechanism in Definition~\ref{def3} dynamically allocates the largest relative weight to the most accurate human labeler 5. As $t$ increases, those inaccurate human labelers 1-4' weights decrease substantially towards 0 to play no effect in RLHF in the end. 

\begin{figure}
    \centering    \includegraphics[width=0.6\linewidth]{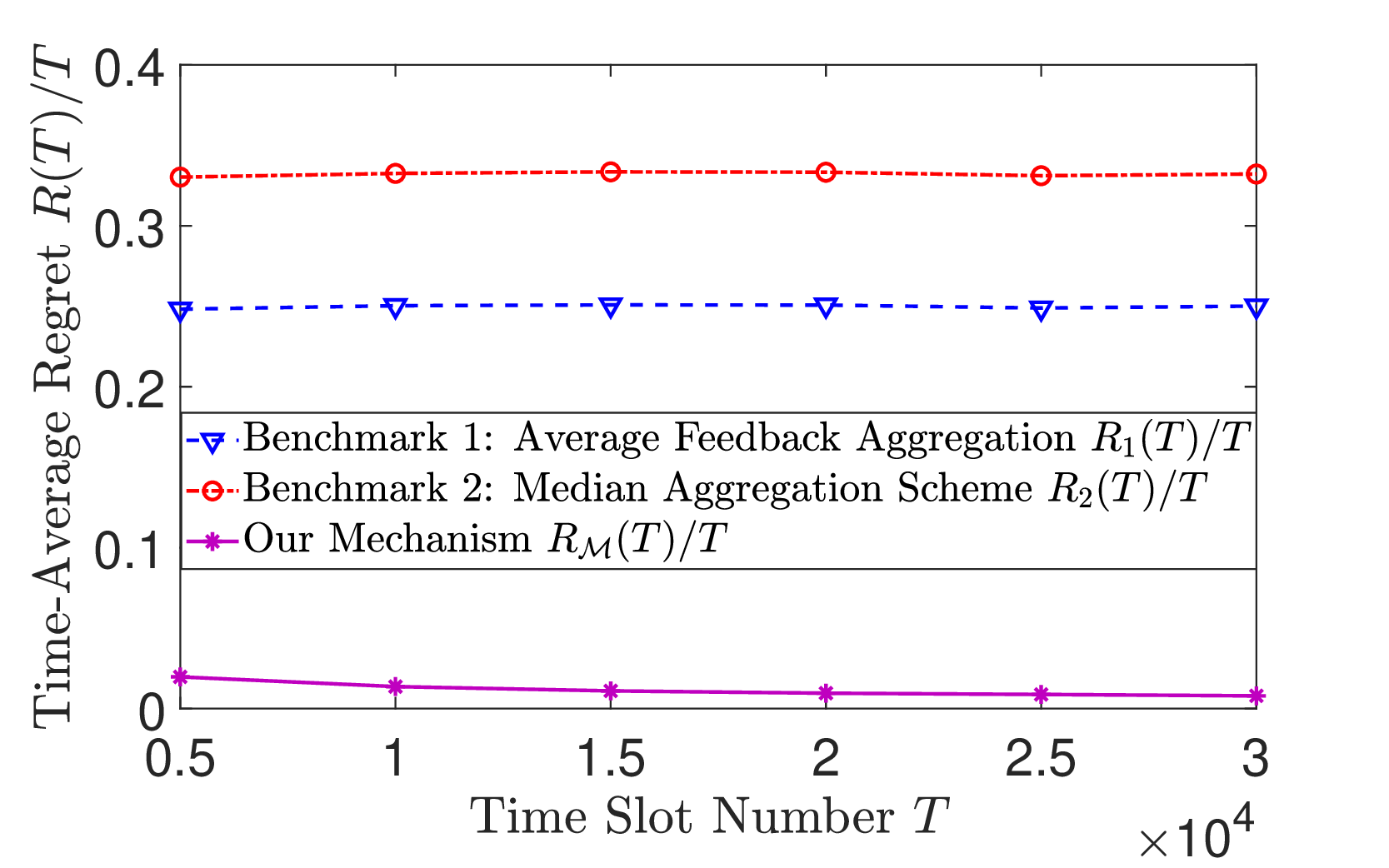}
    \caption{Time-average regrets of benchmarks 1, 2, and our mechanism versus the time slot number $T$, respectively. Here we fix $N=100$ and $m_t=50$.} 
    \label{fig1}
\end{figure}

Figure~\ref{fig1} shows the time-average regrets $R(T)/T$ of benchmarks 1, 2, and our mechanism versus the time slot number $T$. We find that the system's time-average regret is greatly reduced by our mechanism from the two benchmarks. Besides, time-average regrets of both benchmarks 1 and 2 do not decrease with $T$ and are always great than zero, respectively. In constrast, our mechanism's time-average regret decreases with $T$ to approach 0, consistent with Lemmas~\ref{L1}, \ref{L1-1} and Theorem~\ref{Thm1}.

\section{Conclusion}\label{S6}
In this paper, we are the first to study online learning from strategic human
feedback in LLM fine-tuning. We design an efficient truthful mechanism to achieve zero time-average regret, greatly improving from non-vanishing regrets of the average feedback aggregation and median schemes, respectively. Finally, simulation results demonstrate our mechanism's great advantages over the two benchmark schemes.

% ~\newpage


\newpage
\onecolumn
\appendix 

\subsection{Proof of Lemma~\ref{L1}}

We will prove $R_1(T) = \mathcal{O}(T)$ with a possible sequence of human labelers' preferences. In particular, we consider $\mathcal{P}_k(y_{w_j}^t \succ y_{l_j}^t|x_j^t) = p_j^t$ holds for one particular $k \in [N]$ with any $j \in [m_t]$ and $t \in [T]$. For the remaining human labelers, we consider  $(\sum_{i=1,i\ne k}^N \frac{1}{N}\mathcal{\hat{P}}_i(y_{w_j}^t \succ y_{l_j}^t|x_j^t)-\frac{N-1}{N}p_j^t)^2=c_j^t$ for each $i \ne k, i \in [N]$, $j \in [m_t]$ and $t \in [T]$, where $c_j^t \in [\frac{1}{2}, 1]$. Accordingly, we have the best-fixed human labeler in hindsight is $i^* = k$, which brings
\begin{align*}
        \min_{i \in [N]} \sum_{t=1}^T\frac{1}{m_t}\sum_{j=1}^{m_t}\bigg(\mathcal{P}_i(y_{w_j}^t \succ y_{l_j}^t|x_j^t) - p_j^t\bigg)^2 = 0.
\end{align*}
    However, with the system's uniform weight scheme, we have the cumulative aggregation loss over $T$ slots as follows:
    \begin{align*}    &\sum_{t=1}^T\frac{1}{m_t}\sum_{j=1}^{m_t}\bigg(\sum_{i=1}^N \frac{w_i^{t}\mathcal{\hat{P}}_i(y_{w_j}^t \succ y_{l_j}^t|x_j^t)}{\sum_{i'=1}^N w_{i'}^{t}} - p_j^t\bigg)^2\\
     =& \sum_{t=1}^T\frac{1}{m_t}\sum_{j=1}^{m_t}\bigg(\sum_{i=1}^N \frac{\mathcal{\hat{P}}_i(y_{w_j}^t \succ y_{l_j}^t|x_j^t)}{N} - p_j^t\bigg)^2 \\
=& \sum_{t=1}^T\frac{1}{m_t}\sum_{j=1}^{m_t}\bigg(\sum_{i=1,i\ne k}^N \frac{1}{N}\mathcal{\hat{P}}_i(y_{w_j}^t \succ y_{l_j}^t|x_j^t)-\frac{N-1}{N}p_j^t\bigg)^2 \\
=&  \sum_{t=1}^T \frac{1}{m_t}\sum_{j=1}^{m_t} c_j^t = \mathcal{O}(T),
    \end{align*}
    where the last equality holds because each $c_j^t \in [\frac{1}{2}, 1]$ and $\sum_{t=1}^T \frac{1}{m_t}\sum_{j=1}^{m_t} c_j^t$ does not vanish as $T \to \infty$.
    Finally, we have the regret for bechmark 1 of average aggregation as follows:
\begin{align*}
    R_1(T) = \sum_{t=1}^T\frac{1}{m_t}\sum_{j=1}^{m_t}\bigg(\sum_{i=1}^N  
\frac{w_i^{t}\mathcal{\hat{P}}_i(y_{w_j}^t \succ y_{l_j}^t|x_j^t)}{\sum_{i'=1}^N w_{i'}^{t}} - p_j^t\bigg)^2
    - \min_{i \in [N]} \sum_{t=1}^T\frac{1}{m_t}\sum_{j=1}^{m_t}\bigg(\mathcal{P}_i(y_{w_j}^t \succ y_{l_j}^t|x_j^t) - p_j^t\bigg)^2 = \mathcal{O}(T),
\end{align*}
which indicates that $\lim_{T\to\infty}\frac{R_1(T)}{T}> 0$.
We then finish the proof.

\subsection{Proof of Proposition~\ref{L0}}

We want to prove that $\mathcal{\hat{P}}_i^*(y_{w_j}^t \succ y_{l_j}^t|x_j^t)$=$ \mathcal{\hat{P}}(y_{w_j}^t$$\succ $$y_{l_j}^t|x_j^t)$$\in$$[0, 1]$ for $j$$\in$$[m_t]$, $i$$\in$$[N]$ and $t$$\in$$[T]$ is an equilibrium. Note that the system independent determines the aggregation as the median feedback for each prompt $j \in [m_t]$ in each time slot $t\in[T]$, each human labeler $i \in [N]$ aims to maximize each $w_i^t$ to obtain a largest possible accumulative weight overtime. Since the system always commits to the median feedback, given the other $N-1$ human labelers except for $i$ choose $\mathcal{\hat{P}}_k(y_{w_j}^t \succ y_{l_j}^t|x_j^t)$=$ \mathcal{\hat{P}}(y_{w_j}^t$$\succ $$y_{l_j}^t|x_j^t)$, $k \ne i$, $k \in [N]$, the human labeler $i$ feedback any $\mathcal{\hat{P}}_i(y_{w_j}^t \succ y_{l_j}^t|x_j^t)$$\ne$$ \mathcal{\hat{P}}(y_{w_j}^t$$\succ $$y_{l_j}^t|x_j^t)$ will lead to his weight $w_i^t = 0$ because such feedback cannot be the median. Thus, he will feedback consistently as $\mathcal{\hat{P}}_i(y_{w_j}^t \succ y_{l_j}^t|x_j^t)$$=$$ \mathcal{\hat{P}}(y_{w_j}^t$$\succ $$y_{l_j}^t|x_j^t)$ for an equal chance to be the median and will never deviate from this feedback strategy. We then finish the proof.

\subsection{Proof of Lemma~\ref{L1-1}}

We want to prove $R_2(T) = \mathcal{O}(T)$ with a possible sequence of human labelers' preferences. In particular, we consider $\mathcal{P}_k(y_{w_j}^t \succ y_{l_j}^t|x_j^t) = p_j^t$ holds for one particular $k \in [N]$ with any $j \in [m_t]$ and $t \in [T]$. Further, we consider  $(\mathcal{\hat{P}}_{j, k_m}^t-p_j^t)^2=c_j^t$ for $j \in [m_t]$ and $t \in [T]$, where $\mathcal{\hat{P}}_{j, k_m}^t$ denotes the median of human labelers' feedback $\{\mathcal{\hat{P}}_i(y_{w_j}^t \succ y_{l_j}^t|x_j^t)\}_{i=1}^N$ and $c_j^t \in [\frac{1}{2}, 1]$. Accordingly, we have the best-fixed human labeler in hindsight is $i^* = k$, which brings
    \begin{align*}
        \min_{i \in [N]} \sum_{t=1}^T\frac{1}{m_t}\sum_{j=1}^{m_t}\bigg(\mathcal{P}_i(y_{w_j}^t \succ y_{l_j}^t|x_j^t) - p_j^t\bigg)^2 = 0.
    \end{align*}
    However, with the system's median scheme, we have the cumulative aggregation loss over $T$ slots as follows:
    \begin{align*}    
    &\sum_{t=1}^T\frac{1}{m_t}\sum_{j=1}^{m_t}\bigg(\sum_{i=1}^N \frac{w_i^{t}\mathcal{\hat{P}}_i(y_{w_j}^t \succ y_{l_j}^t|x_j^t)}{\sum_{i'=1}^N w_{i'}^{t}} - p_j^t\bigg)^2\\ =&\sum_{t=1}^T\frac{1}{m_t}\sum_{j=1}^{m_t}\bigg(\mathcal{\hat{P}}_{j, k_m}^t - y_j^{t}\bigg)^2 \\
=&  \sum_{t=1}^T \frac{1}{m_t}\sum_{j=1}^{m_t} c_j^t = \mathcal{O}(T),
    \end{align*}
    where the last equality holds because each $c_j^t \in [\frac{1}{2}, 1]$ and $\sum_{t=1}^T \frac{1}{m_t}\sum_{j=1}^{m_t} c_j^t$ does not vanish as $T \to \infty$.
    Finally, we have the regret of the median scheme as follows:
\begin{align*}
    R_2(T) = \sum_{t=1}^T\frac{1}{m_t}\sum_{j=1}^{m_t}\bigg(\sum_{i=1}^N  
\frac{w_i^{t}\mathcal{\hat{P}}_i(y_{w_j}^t \succ y_{l_j}^t|x_j^t)}{\sum_{i'=1}^N w_{i'}^{t}} - p_j^t\bigg)^2
    - \min_{i \in [N]} \sum_{t=1}^T\frac{1}{m_t}\sum_{j=1}^{m_t}\bigg(\mathcal{P}_i(y_{w_j}^t \succ y_{l_j}^t|x_j^t) - p_j^t\bigg)^2 = \mathcal{O}(T).
\end{align*}
We then finish the proof.

\subsection{Proof of Proposition~\ref{L2}}

Note that each human labeler believes that $p_j^t \sim \texttt{Bernoulli}(\mathcal{P}_i(y_{w_j}^t \succ y_{l_j}^t|x_j^t))$, we have expectation on $ w_{i}^{t+1}$ in \eqref{e4} over $p_j^t$ is
    \begin{align*}
        &\mathbb{E}[w_{i}^{t+1}] \\
        =&  w_{i}^{t}\frac{1}{m_t}\sum_{j=1}^{m_t}\bigg[ 1  -\alpha \mathcal{P}_i(y_{w_j}^t \succ y_{l_j}^t|x_j^t)(\mathcal{\hat{P}}_i(y_{w_j}^t \succ y_{l_j}^t|x_j^t)- 1)^2 - \alpha (1-\mathcal{P}_i(y_{w_j}^t \succ y_{l_j}^t|x_j^t))(\mathcal{\hat{P}}_i(y_{w_j}^t \succ y_{l_j}^t|x_j^t)- 0)^2 \bigg] \\
        =& w_{i}^{t}\frac{1}{m_t}\sum_{j=1}^{m_t}\bigg[1 - \alpha (\mathcal{\hat{P}}_i(y_{w_j}^t \succ y_{l_j}^t|x_j^t) - \mathcal{P}_i(y_{w_j}^t \succ y_{l_j}^t|x_j^t))^2 - \alpha (\mathcal{P}_i(y_{w_j}^t \succ y_{l_j}^t|x_j^t) - \mathcal{P}_i^2(y_{w_j}^t \succ y_{l_j}^t|x_j^t)) \bigg],
    \end{align*}
    which is maximized at $\mathcal{\hat{P}}_i^*(y_{w_j}^t \succ y_{l_j}^t|x_j^t) = \mathcal{P}_i(y_{w_j}^t \succ y_{l_j}^t|x_j^t)$. To obtain the largest possible accumulative weight, each human labeler will truthfully feedback his preference in the first time slot and all the following time slots because any deviation will lead to  smaller weights of the next and all the following time slots. We then finish the proof.

\subsection{Proof of Theorem~\ref{Thm1}}

According to Proposition~\ref{L2}, we have $\mathcal{\hat{P}}_i(y_{w_j}^t \succ y_{l_j}^t|x_j^t) = \mathcal{P}_i(y_{w_j}^t \succ y_{l_j}^t|x_j^t)$ for all $j \in [m_t]$, $i \in [N]$ and $t \in [T]$. To derive a lower-bound on $\ln\frac{\sum_{i=1}^N w_{i}^{T+1}}{\sum_{i=1}^N w_{i}^1}$, we have
\begin{align}
  \ln\frac{\sum_{i=1}^N w_{i}^{T+1}}{\sum_{i=1}^N w_{i}^1} =& \ln\bigg( \sum_{i=1}^N w_{i}^{T+1} \bigg) - \ln\bigg( \sum_{i=1}^N w_{i}^1 \bigg) \nonumber \\
  =& \ln \bigg( \sum_{i=1}^N \prod_{t=1}^T ( 1-\alpha\frac{1}{m_t}\sum_{j=1}^{m_t} (\mathcal{P}_i(y_{w_j}^t \succ y_{l_j}^t|x_j^t)- p_j^t)^2 ) \bigg) - \ln N \nonumber \\
  \geq& \ln \bigg( \prod_{t=1}^T ( 1-\alpha\frac{1}{m_t}\sum_{j=1}^{m_t} (\mathcal{P}_{i^*}(y_{w_j}^t \succ y_{l_j}^t|x_j^t)- p_j^t)^2 )  \bigg) - \ln N \nonumber \\
% \end{align}
% \begin{align}
  =& \sum_{t=1}^T \ln \bigg( 1-\alpha \frac{1}{m_t}\sum_{j=1}^{m_t}(\mathcal{P}_{i^*}(y_{w_j}^t \succ y_{l_j}^t|x_j^t)- p_j^t)^2   \bigg) - \ln N \nonumber \\
  \geq& -\alpha\sum_{t=1}^T\frac{1}{m_t}\sum_{j=1}^{m_t} (\mathcal{P}_{i^*}(y_{w_j}^t \succ y_{l_j}^t|x_j^t)- p_j^t)^2 -\alpha^2\sum_{t=1}^T\bigg(\frac{1}{m_t}\sum_{j=1}^{m_t} (\mathcal{P}_{i^*}(y_{w_j}^t \succ y_{l_j}^t|x_j^t)- p_j^t)^2\bigg)^2 - \ln N \nonumber \\
  \geq& -\alpha\sum_{t=1}^T\frac{1}{m_t}\sum_{j=1}^{m_t} \bigg(\mathcal{P}_{i^*}(y_{w_j}^t \succ y_{l_j}^t|x_j^t)- p_j^t\bigg)^2 -\alpha^2 T - \ln N, \label{low'}
\end{align}
where we choose $\alpha < \frac{1}{2}$ and denote $i^*$ as the best human labeler in hindsight. The first and the third inequalities hold due to $0 < \alpha < \frac{1}{2}$ and $0 \leq (\mathcal{P}_{i^*}(y_{w_j}^t \succ y_{l_j}^t|x_j^t)- p_j^t)^2 \leq 1$ for all $i \in [N]$ and $t \in [T]$. The second inequality holds due to $\ln (1 - x) \geq -x -x^2$ for $x \leq \frac{1}{2}$.

To derive an upper-bound on $\ln\frac{\sum_{i=1}^N w_{i}^{t+1}}{\sum_{i=1}^N w_{i}^t}$, we have
\begin{align}
  &\ln\frac{\sum_{i=1}^N w_{i}^{t+1}}{\sum_{i=1}^N w_{i}^t} \nonumber \\
  =& \ln \bigg( \frac{\sum_{i=1}^N w_{i}^t \cdot (1-\alpha\frac{1}{m_t}\sum_{j=1}^{m_t} (\mathcal{P}_i(y_{w_j}^t \succ y_{l_j}^t|x_j^t)- p_j^t)^2)}{\sum_{i'=1}^N w_{i'}^t} \bigg) \nonumber \\
  \leq& \ln \bigg( \frac{\sum_{i=1}^N w_{i}^t \cdot e^{-\alpha\frac{1}{m_t}\sum_{j=1}^{m_t} (\mathcal{P}_i(y_{w_j}^t \succ y_{l_j}^t|x_j^t)- p_j^t)^2}}{\sum_{i'=1}^N w_{i'}^t} \bigg) \nonumber \\
  % \end{align}
  % \begin{align}
  \leq&  -\alpha \frac{1}{m_t}\sum_{j=1}^{m_t}\frac{\sum_{i=1}^N w_{i}^t(\mathcal{P}_i(y_{w_j}^t \succ y_{l_j}^t|x_j^t)- p_j^t)^2}{\sum_{i'=1}^N w_{i'}^t}  + \frac{\alpha^2}{8},  \label{upp'}
\end{align}
where the first inequality holds due to $1-\alpha x \leq e^{-\alpha x}$ for $0 \leq x \leq 1$ and $\alpha > 0$, the second due to Hoeffding's lemma: for a random variable $X = -\frac{1}{m_t}\sum_{j=1}^{m_t}(\mathcal{P}_i(y_{w_j}^t \succ y_{l_j}^t|x_j^t)- p_j^t)^2  \in [-1, 0]$ and $\alpha \in R$, we have
\begin{align*}
    \ln (\mathbf{E}[e^{\alpha X}]) \leq \alpha \mathbf{E}[X] + \frac{\alpha^2(1-0)^2}{8}.
\end{align*}
According to \eqref{upp'}, we have

\begin{align}
    \ln\frac{\sum_{i=1}^N w_{i}^{T+1}}{\sum_{i=1}^N w_{i}^1} &= \ln\bigg(\frac{\sum_{i=1}^N w_{i}^{T+1}}{\sum_{i=1}^N w_{i}^t} \frac{\sum_{i=1}^N w_{i}^t}{\sum_{i=1}^N w_{i}^{t-1}} \cdot\cdots\cdot\frac{\sum_{i=1}^N w_{i}^2}{\sum_{i=1}^N w_{i}^1}\bigg) = \sum_{t=1}^T \ln\frac{\sum_{i=1}^N w_{i}^{T+1}}{\sum_{i=1}^N w_{i}^t} \nonumber \\
% \end{align}
% \begin{align}
    &\leq -\alpha\sum_{t=1}^T\frac{1}{m_t}\sum_{j=1}^{m_t} \frac{\sum_{i=1}^N w_{i}^t(\mathcal{P}_i(y_{w_j}^t \succ y_{l_j}^t|x_j^t)- p_j^t)^2}{\sum_{i'=1}^N w_{i'}^t}  + \frac{\alpha^2 T}{8}. \label{upp''}
\end{align}
According to \eqref{low'} and \eqref{upp''}, we have
\begin{align*}
&-\alpha\sum_{t=1}^T\frac{1}{m_t}\sum_{j=1}^{m_t} \bigg(\mathcal{P}_{i^*}(y_{w_j}^t \succ y_{l_j}^t|x_j^t)- p_j^t\bigg)^2 -\alpha^2 T - \ln N \\ 
\leq& -\alpha\sum_{t=1}^T\frac{1}{m_t}\sum_{j=1}^{m_t} \frac{\sum_{i=1}^N w_{i}^t(\mathcal{P}_i(y_{w_j}^t \succ y_{l_j}^t|x_j^t)- p_j^t)^2}{\sum_{i'=1}^N w_{i'}^t}  + \frac{\alpha^2 T}{8}.
\end{align*}
After re-arranging the above inequalities and dividing $\alpha$ on both sides, we have
\begin{align*}
   \sum_{t=1}^T\frac{1}{m_t}\sum_{j=1}^{m_t} \frac{\sum_{i=1}^N w_{i}^t(\mathcal{P}_i(y_{w_j}^t \succ y_{l_j}^t|x_j^t)- p_j^t)^2}{\sum_{i'=1}^N w_{i'}^t}  - \sum_{t=1}^T\frac{1}{m_t}\sum_{j=1}^{m_t} \bigg(\mathcal{P}_{i^*}(y_{w_j}^t \succ y_{l_j}^t|x_j^t)- p_j^t\bigg)^2 
   \leq \frac{\ln N}{\alpha} + \frac{9T\alpha}{8}.
\end{align*}
Choosing $\alpha = \frac{2}{3}\sqrt{\frac{2\ln N}{T}} < \frac{1}{2}$, we have
\begin{align*}
  \sum_{t=1}^T\frac{1}{m_t}\sum_{j=1}^{m_t} \frac{\sum_{i=1}^N w_{i}^t(\mathcal{P}_i(y_{w_j}^t \succ y_{l_j}^t|x_j^t)- p_j^t)^2}{\sum_{i'=1}^N w_{i'}^t}  - \sum_{t=1}^T\frac{1}{m_t}\sum_{j=1}^{m_t} \bigg(\mathcal{P}_{i^*}(y_{w_j}^t \succ y_{l_j}^t|x_j^t)- p_j^t\bigg)^2 \leq 3\sqrt{\frac{T\ln N}{2}}.
\end{align*}
Finally, we have the regret $R_{\mathcal{M}}(T)$ satisfying
\begin{align*}
    R_{\mathcal{M}}(T) &= \sum_{t=1}^T\frac{1}{m_t}\sum_{j=1}^{m_t}\bigg(\sum_{i=1}^N \frac{w_i^{t}\mathcal{\hat{P}}_i(y_{w_j}^t \succ y_{l_j}^t|x_j^t)}{\sum_{i'=1}^N w_{i'}^{t}} - p_j^t\bigg)^2
    - \sum_{t=1}^T\frac{1}{m_t}\sum_{j=1}^{m_t}\bigg(\mathcal{P}_{i^*}(y_{w_j}^t \succ y_{l_j}^t|x_j^t) - p_j^t\bigg)^2 \\
    &\leq \sum_{t=1}^T\frac{1}{m_t}\sum_{j=1}^{m_t} \frac{\sum_{i=1}^N w_{i}^t(\mathcal{P}_i(y_{w_j}^t \succ y_{l_j}^t|x_j^t)- p_j^t)^2}{\sum_{i'=1}^N w_{i'}^t}  - \sum_{t=1}^T\frac{1}{m_t}\sum_{j=1}^{m_t} \bigg(\mathcal{P}_{i^*}(y_{w_j}^t \succ y_{l_j}^t|x_j^t)- p_j^t\bigg)^2 \\
    &\leq3\sqrt{\frac{T\ln N}{2}} = \mathcal{O}(T^{\frac{1}{2}}),
\end{align*}
where the first inequality holds due to the convexity of the aggregation loss function. We then finish the proof.


\begin{thebibliography}{10}
\providecommand{\url}[1]{#1}
\csname url@samestyle\endcsname
\providecommand{\newblock}{\relax}
\providecommand{\bibinfo}[2]{#2}
\providecommand{\BIBentrySTDinterwordspacing}{\spaceskip=0pt\relax}
\providecommand{\BIBentryALTinterwordstretchfactor}{4}
\providecommand{\BIBentryALTinterwordspacing}{\spaceskip=\fontdimen2\font plus
\BIBentryALTinterwordstretchfactor\fontdimen3\font minus \fontdimen4\font\relax}
\providecommand{\BIBforeignlanguage}[2]{{%
\expandafter\ifx\csname l@#1\endcsname\relax
\typeout{** WARNING: IEEEtran.bst: No hyphenation pattern has been}%
\typeout{** loaded for the language `#1'. Using the pattern for}%
\typeout{** the default language instead.}%
\else
\language=\csname l@#1\endcsname
\fi
#2}}
\providecommand{\BIBdecl}{\relax}
\BIBdecl

\bibitem{touvron2023open}
H.~Touvron, T.~Lavril, G.~Izacard, X.~Martinet, M.~Lachaux, T.~Lacroix, B.~Rozi{\`e}re, N.~Goyal, E.~Hambro, F.~Azhar \emph{et~al.}, ``Open and efficient foundation language models,'' \emph{Preprint at arXiv. https://doi. org/10.48550/arXiv}, vol. 2302, 2023.

\bibitem{kopf2024openassistant}
A.~K{\"o}pf, Y.~Kilcher, D.~von R{\"u}tte, S.~Anagnostidis, Z.~R. Tam, K.~Stevens, A.~Barhoum, D.~Nguyen, O.~Stanley, R.~Nagyfi \emph{et~al.}, ``Openassistant conversations-democratizing large language model alignment,'' \emph{Advances in Neural Information Processing Systems}, vol.~36, 2024.

\bibitem{sun2024mechanism}
H.~Sun, Y.~Chen, S.~Wang, W.~Chen, and X.~Deng, ``Mechanism design for llm fine-tuning with multiple reward models,'' \emph{arXiv preprint arXiv:2405.16276}, 2024.

\bibitem{christiano2017deep}
P.~F. Christiano, J.~Leike, T.~Brown, M.~Martic, S.~Legg, and D.~Amodei, ``Deep reinforcement learning from human preferences,'' \emph{Advances in neural information processing systems}, vol.~30, 2017.

\bibitem{ouyang2022training}
L.~Ouyang, J.~Wu, X.~Jiang, D.~Almeida, C.~Wainwright, P.~Mishkin, C.~Zhang, S.~Agarwal, K.~Slama, A.~Ray \emph{et~al.}, ``Training language models to follow instructions with human feedback,'' \emph{Advances in neural information processing systems}, vol.~35, pp. 27\,730--27\,744, 2022.

\bibitem{bai2022training}
Y.~Bai, A.~Jones, K.~Ndousse, A.~Askell, A.~Chen, N.~DasSarma, D.~Drain, S.~Fort, D.~Ganguli, T.~Henighan \emph{et~al.}, ``Training a helpful and harmless assistant with reinforcement learning from human feedback. corr, abs/2204.05862, 2022a. doi: 10.48550,'' \emph{arXiv preprint arXiv.2204.05862}, 2022.

\bibitem{touvron2023llama}
H.~Touvron, L.~Martin, K.~Stone, P.~Albert, A.~Almahairi, Y.~Babaei, N.~Bashlykov, S.~Batra, P.~Bhargava, S.~Bhosale \emph{et~al.}, ``Llama 2: Open foundation and fine-tuned chat models,'' \emph{arXiv preprint arXiv:2307.09288}, 2023.

\bibitem{xiong2024iterative}
W.~Xiong, H.~Dong, C.~Ye, Z.~Wang, H.~Zhong, H.~Ji, N.~Jiang, and T.~Zhang, ``Iterative preference learning from human feedback: Bridging theory and practice for rlhf under kl-constraint,'' in \emph{Forty-first International Conference on Machine Learning}, 2024.

\bibitem{soumalias2024truthful}
E.~Soumalias, M.~J. Curry, and S.~Seuken, ``Truthful aggregation of llms with an application to online advertising,'' \emph{arXiv preprint arXiv:2405.05905}, 2024.

\bibitem{conitzer2024social}
V.~Conitzer, R.~Freedman, J.~Heitzig, W.~H. Holliday, B.~M. Jacobs, N.~Lambert, M.~Moss{\'e}, E.~Pacuit, S.~Russell, H.~Schoelkopf \emph{et~al.}, ``Social choice for ai alignment: Dealing with diverse human feedback,'' \emph{arXiv preprint arXiv:2404.10271}, 2024.

\bibitem{roughgarden2017online}
T.~Roughgarden and O.~Schrijvers, ``Online prediction with selfish experts,'' \emph{Advances in Neural Information Processing Systems}, vol.~30, 2017.

\bibitem{rafailov2024direct}
R.~Rafailov, A.~Sharma, E.~Mitchell, C.~D. Manning, S.~Ermon, and C.~Finn, ``Direct preference optimization: Your language model is secretly a reward model,'' \emph{Advances in Neural Information Processing Systems}, vol.~36, 2024.

\bibitem{xie2024exploratory}
T.~Xie, D.~J. Foster, A.~Krishnamurthy, C.~Rosset, A.~Awadallah, and A.~Rakhlin, ``Exploratory preference optimization: Harnessing implicit q*-approximation for sample-efficient rlhf,'' \emph{arXiv preprint arXiv:2405.21046}, 2024.

\bibitem{cen2024value}
S.~Cen, J.~Mei, K.~Goshvadi, H.~Dai, T.~Yang, S.~Yang, D.~Schuurmans, Y.~Chi, and B.~Dai, ``Value-incentivized preference optimization: A unified approach to online and offline rlhf,'' \emph{arXiv preprint arXiv:2405.19320}, 2024.

\bibitem{zhang2024self}
S.~Zhang, D.~Yu, H.~Sharma, Z.~Yang, S.~Wang, H.~Hassan, and Z.~Wang, ``Self-exploring language models: Active preference elicitation for online alignment,'' \emph{arXiv preprint arXiv:2405.19332}, 2024.

\bibitem{chen2024optune}
L.~Chen, J.~Chen, C.~Liu, J.~Kirchenbauer, D.~Soselia, C.~Zhu, T.~Goldstein, T.~Zhou, and H.~Huang, ``Optune: Efficient online preference tuning,'' \emph{arXiv preprint arXiv:2406.07657}, 2024.

\bibitem{park2024rlhf}
C.~Park, M.~Liu, D.~Kong, K.~Zhang, and A.~E. Ozdaglar, ``Rlhf from heterogeneous feedback via personalization and preference aggregation,'' in \emph{ICML 2024 Workshop on Theoretical Foundations of Foundation Models}, 2024.

\bibitem{dubey2024auctions}
K.~A. Dubey, Z.~Feng, R.~Kidambi, A.~Mehta, and D.~Wang, ``Auctions with llm summaries,'' \emph{arXiv preprint arXiv:2404.08126}, 2024.

\bibitem{asadi2022collaborative}
M.~Asadi, A.~Bellet, O.-A. Maillard, and M.~Tommasi, ``Collaborative algorithms for online personalized mean estimation,'' \emph{Transactions on Machine Learning Research Journal}, 2022.

\bibitem{chen2024mechanism}
Y.~Chen, J.~Zhu, and K.~Kandasamy, ``Mechanism design for collaborative normal mean estimation,'' \emph{Advances in Neural Information Processing Systems}, vol.~36, 2024.

\bibitem{li2024strategyproof}
J.~Li, M.~Li, and H.~Chan, ``Strategyproof mechanisms for group-fair obnoxious facility location problems,'' in \emph{Proceedings of the AAAI Conference on Artificial Intelligence}, vol.~38, no.~9, 2024, pp. 9832--9839.

\bibitem{wang2024positive}
Y.~Wang, H.~Zhou, and M.~Li, ``Positive intra-group externalities in facility location,'' in \emph{Proceedings of the 23rd International Conference on Autonomous Agents and Multiagent Systems}, 2024, pp. 1883--1891.

\bibitem{pacchiano2021dueling}
A.~Pacchiano, A.~Saha, and J.~Lee, ``Dueling rl: reinforcement learning with trajectory preferences,'' \emph{arXiv preprint arXiv:2111.04850}, 2021.

\bibitem{chen2022human}
X.~Chen, H.~Zhong, Z.~Yang, Z.~Wang, and L.~Wang, ``Human-in-the-loop: Provably efficient preference-based reinforcement learning with general function approximation,'' in \emph{International Conference on Machine Learning}.\hskip 1em plus 0.5em minus 0.4em\relax PMLR, 2022, pp. 3773--3793.

\bibitem{zhong2024dpo}
H.~Zhong, G.~Feng, W.~Xiong, L.~Zhao, D.~He, J.~Bian, and L.~Wang, ``Dpo meets ppo: Reinforced token optimization for rlhf,'' \emph{arXiv preprint arXiv:2404.18922}, 2024.

\bibitem{frongillo2021efficient}
R.~Frongillo, R.~Gomez, A.~Thilagar, and B.~Waggoner, ``Efficient competitions and online learning with strategic forecasters,'' in \emph{Proceedings of the 22nd ACM Conference on Economics and Computation}, 2021, pp. 479--496.

\bibitem{freeman2020no}
R.~Freeman, D.~Pennock, C.~Podimata, and J.~W. Vaughan, ``No-regret and incentive-compatible online learning,'' in \emph{International Conference on Machine Learning}.\hskip 1em plus 0.5em minus 0.4em\relax PMLR, 2020, pp. 3270--3279.

\bibitem{havrilla2023trlx}
A.~Havrilla, M.~Zhuravinskyi, D.~Phung, A.~Tiwari, J.~Tow, S.~Biderman, Q.~Anthony, and L.~Castricato, ``trlx: A framework for large scale reinforcement learning from human feedback,'' in \emph{Proceedings of the 2023 Conference on Empirical Methods in Natural Language Processing}, 2023, pp. 8578--8595.

\bibitem{volske2017tl}
M.~V{\"o}lske, M.~Potthast, S.~Syed, and B.~Stein, ``Tl; dr: Mining reddit to learn automatic summarization,'' in \emph{Proceedings of the Workshop on New Frontiers in Summarization}, 2017, pp. 59--63.

\end{thebibliography}
\end{document}